\documentclass[sigplan,10pt,nonacm]{acmart}
\AtBeginDocument{%
  }

\usepackage{algorithmicx}
\usepackage{algorithm}
\usepackage{xspace}
\usepackage{booktabs}
\usepackage[table]{xcolor}
\definecolor{bestbg}{HTML}{D9EAD3}
\newcommand{\best}[1]{\cellcolor{bestbg}\textbf{#1}}
\newcommand{\ccd}{\textsc{Concord}\xspace}

\begin{document}
\title{When Agent Context Goes Stale: Incoherence in Volatile Agent Context}
\author{Yingying Liu}
\authornote{Yingying Liu and Junzhou Fang contributed equally to this work.}
\affiliation{%
    \institution{The University of Hong Kong}
    \city{Hong Kong SAR}
    \country{China}
}

\author{Junzhou Fang}
\authornotemark[1]
\affiliation{%
    \institution{The University of Hong Kong}
    \city{Hong Kong SAR}
    \country{China}
}

\author{Chenxiong Qian}
\authornote{Corresponding author}
\affiliation{%
    \institution{The University of Hong Kong}
    \city{Hong Kong SAR}
    \country{China}
}

\begin{abstract}
Modern agents increasingly ground their reasoning in observations returned by tools, such as file contents read from a workspace. 
However, the data sources underlying these observations may later be modified by users, other agents, or external tools, while the model retains only the stale content in its context window. 
Existing agent runtimes provide little support for notifying the model that a previously observed fact has become stale, causing agents to reuse outdated observations and make incorrect claims about the current workspace state.

We propose \ccd, a context coherence framework that maintains the consistency between tool observations in agent context and the mutable sources from which they were derived.
\ccd links each observation to its source, detects source changes, and uses configurable handling policies to update, annotate, or suppress stale context before reuse.
\ccd is applicable across different agent runtimes and external resources, and can be easily extended to new runtime-resource settings.
We implement \ccd as a general framework, and instantiate a concrete use case to assess its effectiveness.
We construct ConcordBench, where previously observed file contents become stale after subsequent edits. 
Across three evaluated frontier models, \ccd produces answers consistent with the restored workspace state in all evaluated cases under these constructed conditions, matching the oracle on recover count for this benchmark, while using 46.4\% fewer tokens than the strongest non-oracle baseline. Our code is available \href{https://github.com/FFFFFFFANG1/Concord}{\underline{here}}.
\end{abstract}

\maketitle
\section{Introduction}
AI agents are increasingly used as tool-using systems that interact with external resources during task execution. 
Modern agent runtimes, such as OpenClaw~\cite{openclaw2026context,openclaw2026tools}, 
OpenCode~\cite{opencode2026mcp,opencode2026github}, and
OpenHands~\cite{wang2025openhands}
support built-in tools and extensions based on the Model Context Protocol (MCP)~\cite{anthropic2024mcp} for reading files, executing commands, querying services, and accessing remote resources.
As a result, an agent's context is no longer limited to the initial user prompt or multi-turn dialog history, but also includes tool observations produced during execution.
These tool observations are typically stored as structured execution entries and later projected into LLM messages before each model call. 

Tool observations, however, are not live views of external resources. 
They are snapshots produced at a particular point in the agent’s execution, while the resources they describe may continue to change outside the agent’s view. 
For example, in multi-agent systems~\cite{hong2024metagpt}, one agent may update a file that another agent has already read; a developer may manually revise a configuration during an agent run; or a database queried through an MCP tool may be modified by an external transaction.
If the agent runtime preserves earlier tool results as reusable context without tracking whether their source data has changed, stale observations may continue to be presented to the model as current facts, leading to incorrect actions or hallucinated claims about the environment.

Existing work addresses related consistency problems at either the workspace or retrieval level.
Workspace-level systems reduce conflicts through write validation, isolated workspaces, synchronization, or snapshot reconstruction~\cite{liu2026multi,geng2026effective,ren2026lemonharness,pugachev2025codecrdt,yu2026infiagent}.
Memory and retrieval systems use freshness-aware updates, conflict detection, provenance, or temporal supersession to decide which stored or retrieved evidence should be used for generation~\cite{hao2026self,wang2026conflictrag,margalit2026governed}.
These approaches show the importance of freshness and consistency, but they do not repair tool observations after those observations have already been materialized inside an agent's private prompt context.
Consequently, obsolete observations can remain in later prompts and continue to shape model reasoning after their sources have changed.

\begin{figure}[t]
  \centering
  \includegraphics[width=\linewidth]{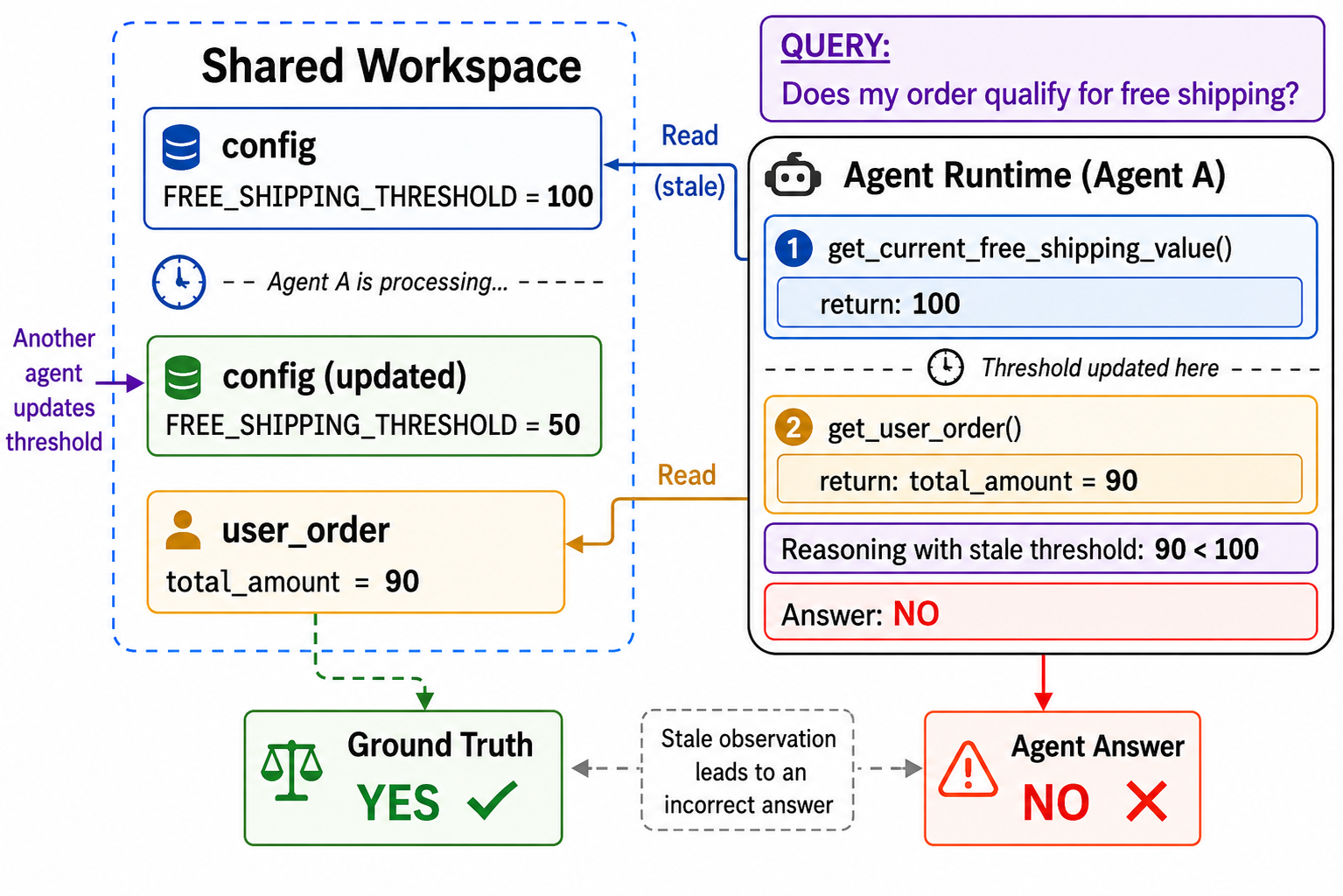}
  \captionsetup{font=footnotesize}
  \caption{A stale-observation failure in an agent-maintained workspace. After the agent records an old free-shipping threshold in its context, the configuration file is edited outside its view. The agent later answers from the stale observation, producing an answer that contradicts the current workspace state.}
  \label{fig:stale-observation-example}
\end{figure}
\textbf{Our approach.}
To address this gap, we propose \ccd, a context coherence framework that maintains the consistency between tool observations in agent context and the mutable sources from which they were derived.
At its core, \ccd uses a runtime- and source-agnostic Context Manager, which tracks tool observations independently of any particular agent framework or external resource.
Runtime Adapters connect the manager to different agent runtimes, while Source Adapters capture changes from different mutable resources.
\ccd is therefore easy to extend to new runtimes, tools, and resource types.

We make the following contributions:
\begin{itemize}
    \item We identify and formally define incoherence in volatile agent context as a coherence problem in tool-using agents, where context entries may lose consistency with the mutable external sources.
    \item We introduce \ccd, a runtime- and resource-agnostic framework that tracks source-linked context blocks, detects source changes, and reconciles stale observations through configurable policies.
    \item We instantiate \ccd for local file resources and evaluate it on a HotpotQA-derived race benchmark across three frontier language models. Under the constructed race conditions in the 40-task ConcordBench-Pioneer, \ccd achieves a recover count of 40/40 for each evaluated model, matching the oracle recover-count upper bound on this benchmark while using fewer tokens than the evaluated non-oracle baselines.
\end{itemize}
\section{Motivation}
\subsection{Example: Stale Context after Workspace Changes}
As shown in~\autoref{fig:stale-observation-example}, we illustrate the problem with a controlled agentic coding workflow built using the OpenHands SDK~\cite{wang2025openhands}. 
The agent is asked to maintain a shop project, where free shipping applies when an order total is greater than or equal to a configurable threshold. 
Initially, the agent reads the project files and observes that \texttt{FREE\_SHIPPING\_THRESHOLD} is set to 100. 
The runtime then appends the file-reading tool return to the agent's conversation context.
After that, the underlying configuration file is modified outside the agent's view, lowering the threshold from 100 to 50. 
When the agent finally answers whether an order totaling 60 qualifies for free shipping, it answers ``no'' based on the stale threshold value of 100. 
However, the current workspace state implies the opposite answer: since the threshold is now 50, the order should qualify for free shipping.
This example highlights a key limitation of current agent runtimes: tool observations can remain in the conversation history even after the underlying workspace state has changed, causing the agent to reason from stale context.

\subsection{Why worth a general framework?}
Volatile tool observations are not specific to files.
They arise whenever an agent observes mutable external state through a tool.
In production settings, such state may come from database records exposed by Postgres servers~\cite{postgres_mcp}, issue and pull-request status from a GitHub server~\cite{github_mcp_server}, or documents in a Slack workspace~\cite{slack_mcp_server}.
Each tool invocation gives the agent a snapshot of that state, and the snapshot may become outdated as soon as the underlying resource is modified.
A resource-specific repair mechanism is not enough.

Our proposed abstraction treats tool observations as mutable context entries rather than immutable transcript items.
Each entry is associated with a runtime-managed identifier and source metadata, such as the external resource it was derived from and the relevant content range. The runtime can then maintain an index from mutable sources to context entries and update the corresponding entries when source
changes are observed. 
In this way, context validity is handled through a general runtime layer, while tool-specific logic is limited to registering observations and reporting source mutations.

\section{\ccd: A Context Coherence Framework for Volatile Agent Context}
In this section, we present \textit{\ccd}, a context coherence framework that maintains the consistency between tool observations in agent context and the mutable sources from which they were derived.
We first formalize the incoherence in volatile agent context and then describe \ccd's architecture, focusing on its core data structures and data paths.

\subsection{Problem Formulation}
We model an agent run as a sequence of steps.
Before step \(t\), the model-visible context contains the prompt and the previous action-observation history:
\[
C_A^t = (p,\ a_1,\ o_1,\ \ldots,\ a_{t-1},\ o_{t-1}),
\]
where \(p\) is the prompt, \(a_i\) is the tool action issued at step \(i\), and \(o_i\) is the corresponding tool observation returned to the agent.

Let \(R\) denote an external resource, if an action \(a_i\) reads\footnote{We use read as an abstract notation for any observation-producing access to an external resource, including file reads, database queries, or MCP resource requests.} \(R\) at step \(i\), the agent records
\[
o_i = \mathrm{read}_i(R),
\]
where \(\mathrm{read}_i(R)\) denotes the content returned by reading \(R\) at step \(i\).
Because \(R\) is mutable, a later read may return different content:
\[
\mathrm{read}_j(R) \neq \mathrm{read}_i(R), \qquad j > i.
\]

We say that observation \(o_i\) is \textbf{stale} at step \(j\) if it remains in the agent context but no longer matches the current result of reading the same resource:
\[
\mathrm{stale}(o_i,j) \iff o_i \in C_A^j
\land
o_i \neq \mathrm{read}_j(R).
\]
We define \textbf{incoherence in volatile agent context} as the condition in which the model-visible context contains at least one stale observation:
\[
\mathrm{volatile}(C_A^j) \iff
\exists o_i \in C_A^j:\ \mathrm{stale}(o_i,j).
\]

A stale-observation failure occurs when the agent treats \(o_i\) as current evidence and produces an output that is inconsistent with \(\mathrm{read}_j(R)\).
The research problem is therefore to efficiently detect and reconcile stale observations before they are reused in later model calls.

\begin{figure}[t]
  \centering
  \includegraphics[width=\linewidth]{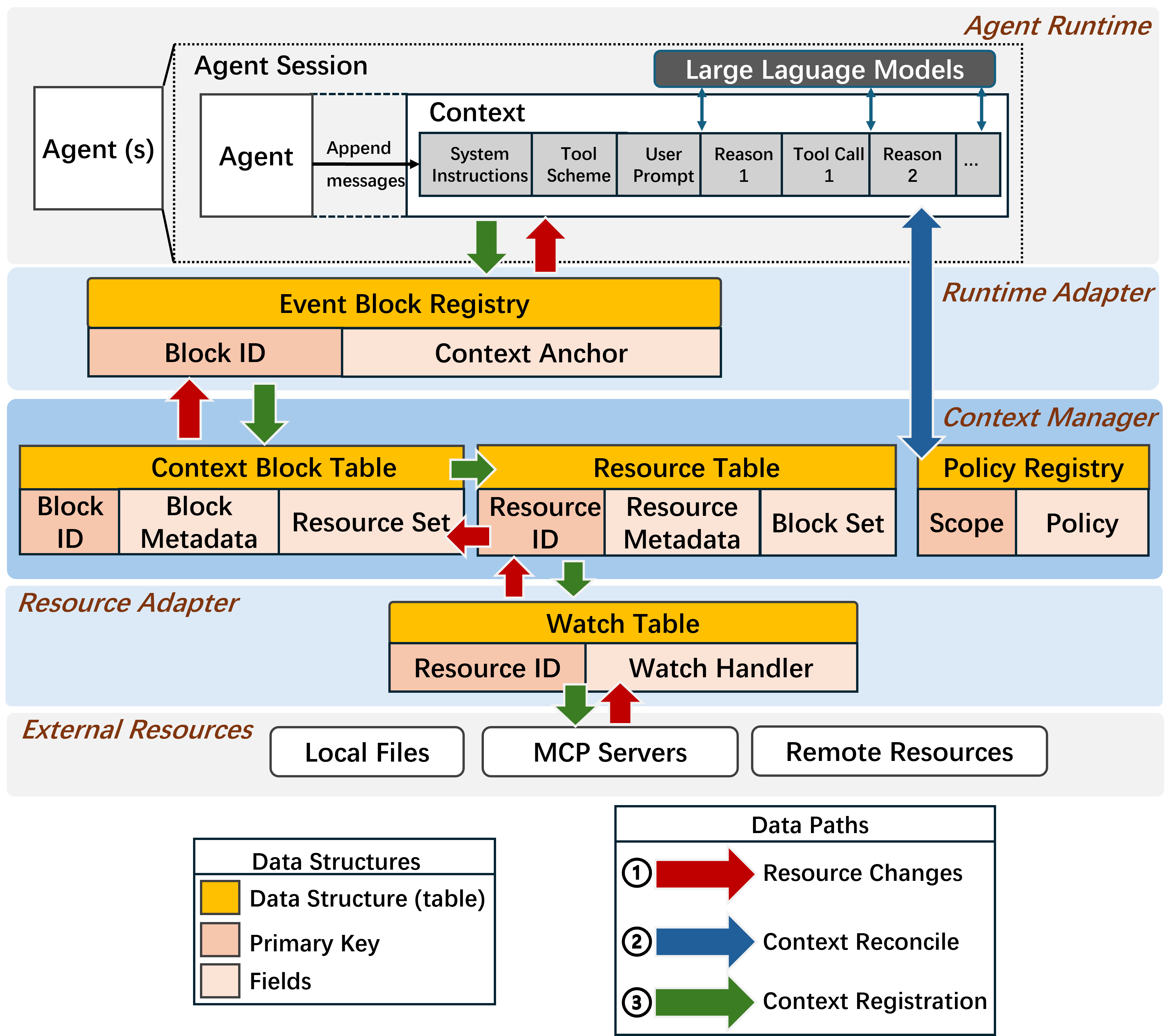}
  \captionsetup{font=footnotesize}
  \caption{Architecture overview of \ccd. A runtime- and source-agnostic \textit{Context Manager} tracks tool observations as source-linked context blocks, connected to the agent runtime and to external resources through a \textit{Runtime Adapter} and a \textit{Resource Adapter}. Colored arrows denote the three data paths: block registration (green), source-change invalidation (red), and policy-driven context reconciliation (blue).}
  \label{fig:overview}
\end{figure}
\subsection{\ccd Architecture}
We address the aforementioned problem by introducing a context layer between the agent runtime and external resources.
%
As shown in~\autoref{fig:overview}, \ccd is an extensible context-coherence framework organized around three components: a runtime adapter layer, a source-agnostic context manager, and a resource adapter layer.

A runtime adapter binds \ccd to a concrete agent runtime.
It observes tool results as they are inserted into the model-visible context \(C_A^t\), maps the corresponding observations \(o_i\) to source-linked context blocks, and records an anchor that identifies where each block appears in \(C_A^t\).
This boundary is necessary because different agent harnesses may store execution histories, tool observations, and LLM-facing messages in different internal representations.

The context manager is the core of the design.
It maintains source-linked context blocks, tracks the relationship between blocks and mutable sources, records validity status, and applies reconciliation policies when sources change.
Importantly, the context manager does not assume a specific runtime or resource type.
It only requires that an observation can be associated with one or more source identifiers, and that source changes can be reported through a common event interface.

The resource adapter connects the context manager to external mutable sources.
It normalizes source identifiers, obtains source versions, determines whether a source change overlaps with a recorded observation, and captures source-change events through hooks or polling.
For example, a file adapter may track absolute paths, file versions, and line ranges, while an MCP or database adapter may track resource keys, timestamps, or version numbers.
Thus, supporting a new resource type only requires implementing a new resource adapter, rather than changing the core context manager.

\subsection{Tracking and Reconciling Context Blocks}
Our context manager maintains context coherence through three steps: registration, invalidation, and reconciliation.

\noindent \textbf{Context registration.}
When a tool observation is inserted into the agent context, the runtime adapter registers it as a source-linked context block.
The block is assigned a stable block identifier and associated with the source metadata from which it was derived, including the source identifier, observed version, and relevant selector or range.
The context manager then records this relation in both directions: from the block to its source set, and from each source to the blocks that depend on it.
This bidirectional mapping is necessary because later updates arrive from the source side, while repairs must be applied to the corresponding context entries.
The runtime adapter also records a context anchor for each block, allowing the repaired block to be written back to the runtime-specific conversation state.

\noindent \textbf{Resource changes.}
When an external source changes, the resource adapter converts the resource-specific signal into a source-change event.
For example, a file adapter may report a modified path, a new version stamp, and the changed range.
The context manager uses the source-to-block mapping to locate all context blocks that may be affected, including blocks held by different agents.
It then applies the block's validity descriptor to decide whether the change actually invalidates the recorded observation.
In the file case, this can be implemented as range-overlap checking: an edit only invalidates a block if the changed range overlaps with the range observed by that block.
This avoids treating every source update as a whole-resource invalidation.

\noindent \textbf{Context reconciliation.}
After affected blocks are marked stale, the context manager reconciles them according to their repair policy.
A policy may refresh the block by re-reading the source, annotate the block with a stale-context warning, suppress it from future prompts, or force the agent to re-read the source before continuing.

\subsection{Extensible Adapter Interface and Policy Configuration}
In order to be configured for different agent runtimes, resource types, and reconciliation requirements, \ccd exposes three extension points.

\noindent \textbf{Runtime extension.}
A runtime adapter defines the boundary between \ccd and an agent runtime.
Agent runtimes may organize execution history, context entries, and LLM-facing messages in different ways.
\ccd only requires the adapter to expose where tool observations enter the context and how repaired blocks can be reflected back into the model-visible state.
Concretely, the adapter identifies tool results that should be registered, records a context anchor for each managed block, and renders repaired blocks back into the runtime's context representation.
This keeps the context manager independent of the runtime's internal implementation.

\noindent \textbf{Resource extension.}
A resource adapter defines how \ccd observes mutable external sources.
It specifies how source identifiers are normalized, how versions are obtained, how selected content is read, and how a source update is compared with a recorded observation.
The key resource-specific component is the watch handle, which detects source changes and reports them to the context manager.
Different deployments may choose different watch handles.
For example, if a local file may be modified by arbitrary applications, the watch handle may need a system-level interception point, such as file-system notifications or instrumentation around write-related system calls, to observe updates outside the agent runtime.
If the deployment only needs to track files modified by agents within a managed workspace, a narrower and cheaper handle can hook into the runtime's file-write tool or workspace-mutation path.
For remote resources, such as MCP servers or database-backed tools, the watch handle may rely on polling, version checks, server-side change notifications, or resource-specific subscription mechanisms.
Therefore, each resource adapter defines a watch scope that matches the deployment's required consistency guaranty.
\ccd then associates registered observations with the corresponding watch handle and reference-counts the handle across dependent blocks.

\noindent \textbf{Policy configuration.}
When a block is marked stale, the context manager resolves a configurable repair policy from the policy registry.
Policies can be configured globally, per resource type, per source, or per block, and may also depend on the severity of the detected change.
For example, if the refreshed source content is substantially different from the recorded observation, the policy may treat the old block as invalid and annotate, suppress, or require an explicit re-read.
If the change is small and localized, the policy may directly refresh or patch the block's model-visible payload.
An eager refresh policy improves freshness by immediately re-reading the source, but may introduce additional latency or tool calls.
This makes the correctness-overhead tradeoff explicit: deployments can choose stronger freshness guarantees at higher runtime cost, or cheaper policies with weaker guarantees.
\section{Implementation and Evaluation}
We implement \ccd as a general framework for keeping agent context consistent with mutable external resources.
To assess its effectiveness, we instantiate and evaluate one concrete use case: local files as the external resource.

\subsection{Use Case Implementation}
To apply \ccd to a concrete setting, developers only need to implement adapters that connect a specific agent runtime and a specific external resource to the framework.
In this use case, we implement a runtime adapter for the agent execution environment and a file-resource adapter for monitoring and re-reading local files.
The file adapter treats each workspace path as a versioned source whose version is the SHA-256 of file contents, and uses \texttt{apply\_patch} as the watch handle: any write that changes the content hash looks up tables in the context manager and runs reconciliation.
For range observations, an anchored locator relocates the recorded span in the updated file and refreshes the block payload when relocation succeeds; otherwise the block is marked invalid.
Whole-file structured observations are recomputed from the new content.

For the configurable reconciliation policy, we compare the old observation text with the refreshed content under a locality threshold: if the edit is local (same token length with at most $\max(2, \lfloor len/5 \rfloor)$ token substitutions, or sequence similarity $\ge 0.78$), the agent-visible block is patched in place and kept valid; if the drift is larger, the block is left non-valid, omitted from prompt projection, and accompanied by a refresh notice that forces an explicit re-read.

\subsection{Evaluation Setup}
\begin{figure}[t]
  \centering
  \includegraphics[width=\linewidth]{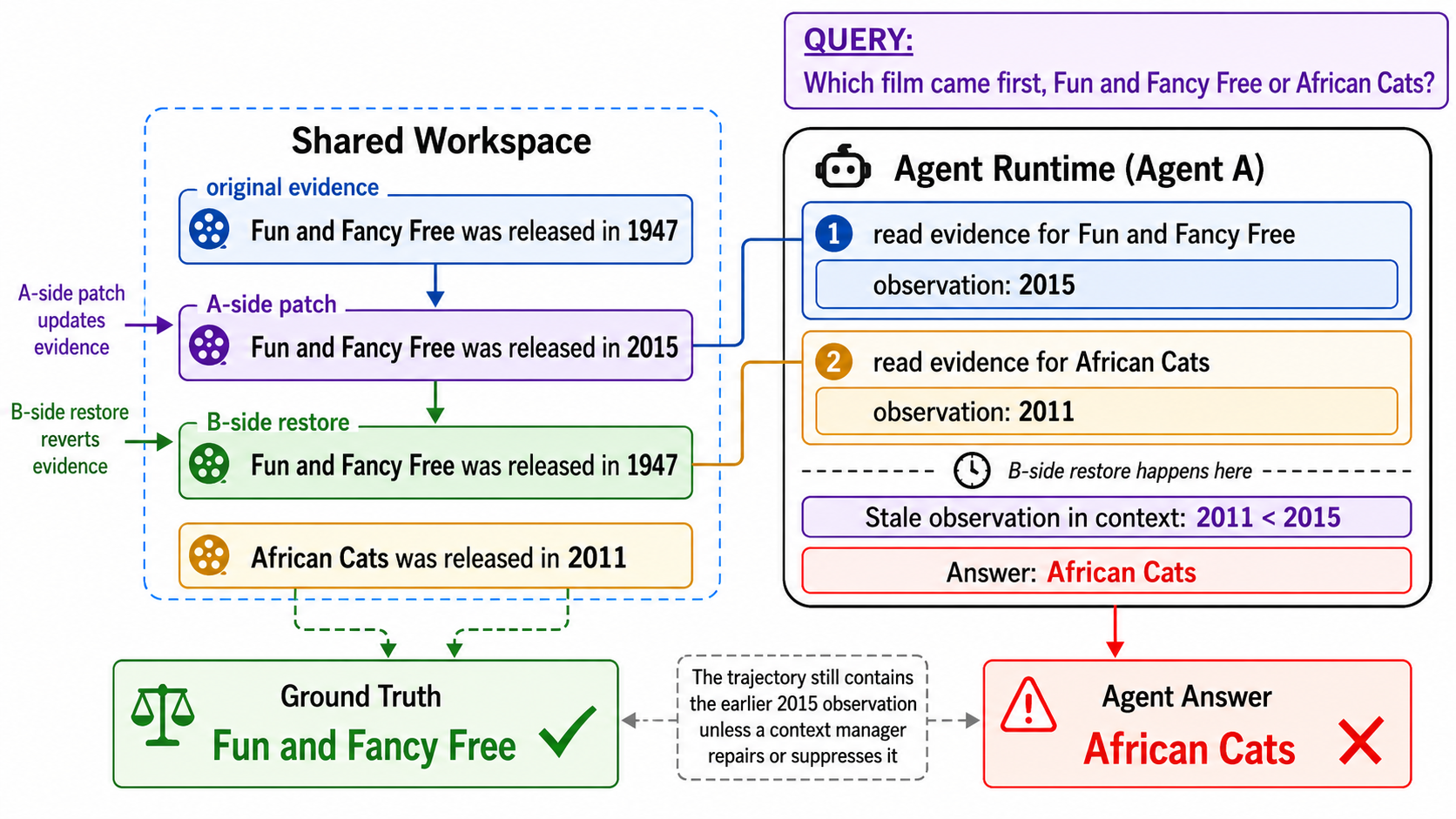}
  \captionsetup{font=footnotesize}
  \caption{Example patch-and-restore race in \textsc{ConcordBench-Pioneer}. The A-side patch makes the agent observe mutated evidence, while the B-side restore reverts the workspace; the agent fails if the stale observation remains in its trajectory.}
  \label{fig:bench-example}
\end{figure}

\noindent\textbf{Benchmark.}
We evaluate on \textsc{ConcordBench-Pioneer}, a 40-task pioneer version of our stale-observation benchmark derived from HotpotQA~\cite{yang2018hotpotqa} multi-hop subset.
Each HotpotQA task contains a question, a gold answer, a set of context documents, and \texttt{supporting\_facts} needed to answer the question.
We materialize the context documents as a local workspace.
During construction, for each task, we select one or more supporting facts, and create a patch for each selected fact by replacing the original evidence sentence with a plausible mutated sentence.
The mutation changes the factual content of the evidence, such as an entity, date, number, or comparison relation, so that an agent relying on the mutated observation is likely to derive an answer different from the original gold answer.
At evaluation time, each task starts from the mutated workspace \(W_{\mathrm{orig}}\) and a patch set \(P\).
Before the agent runs, the benchmark applies the mutated patch and changes the workspace into a mutated workspace:
\[
W_{\mathrm{mut}} = \mathrm{mutate}(W_{\mathrm{orig}}, P).
\]
The agent then interacts with \(W_{\mathrm{mut}}\) using tools.
When the agent reads a patched evidence span, the mutated text is returned by the tool and is recorded in the agent trajectory as a tool observation.
Immediately after that tool observation has entered the trajectory, the race controller applies the B-side restore:
\[
W_{\mathrm{orig}} = \mathrm{restore}(W_{\mathrm{mut}}, P).
\]
Thus, the workspace is again consistent with the original gold answer, but the agent trajectory may still contain the earlier mutated observation.
The final answer is always graded against the original HotpotQA gold answer. Figure \ref{fig:bench-example} shows one evaluation runtime example.

\noindent\textbf{Agent harness.}
The default tool set is limited to \texttt{grep}, \texttt{glob}, \texttt{read\_file}, and \texttt{answer}, excluding shell access and tests so that the benchmark measures stale-context reuse in the trajectory rather than independent environment inspection.
The race controller primarily restores a patch after \texttt{read\_file} returns the patched source; it also restores grep-hit sources if the agent moves on without reading them, and restores remaining patches before premature answer submission.

\noindent\textbf{Baselines.}
We compare \ccd with three baselines, plus an oracle upper bound:
\begin{itemize}
    \item \emph{Naive}: it leaves all prior tool observations unchanged in the context.
    \item \emph{Alert}: it appends \textcolor{blue!55!black}{\texttt{[alert] path is modified}} to the next tool output after a restore (and to the last tool message if the agent tries to answer early), leaving correction to the agent.
    \item \emph{Diff}: it adds a \texttt{diff} tool that reports restored patches as \textcolor{blue!55!black}{\texttt{- mutated / + original}} and blocks \texttt{answer} until the agent has seen a fresh diff for the current workspace generation.
    \item \emph{Oracle}: it runs the same agent and tools on the original workspace with no race injection, serving as an upper bound that isolates HotpotQA difficulty from stale-observation failures.
\end{itemize}

\noindent\textbf{Evaluation metrics.}
We report three evaluation metrics:
\begin{itemize}
    \item \emph{Recover count} is the number of tasks in which the final answer matches the original gold answer after the source has been restored; it is our primary measure of whether the system avoids stale-context hallucination.
    \item \emph{Average steps} measures the number of agent events required to finish the task, where fewer steps indicate lower recovery overhead.
    \item \emph{Token consumption} measures total token usage and captures the prompt-level cost of recovery.
\end{itemize}

\begin{table}[t]
\centering
\small
\setlength{\tabcolsep}{8pt}
\renewcommand{\arraystretch}{1.05}

\caption{Observed results on the 40-task HotpotQA-derived subset. Percentages are empirical recovery rates for this finite benchmark; $\dagger$ denotes the no-race oracle upper-bound setting.}
\vspace{-10pt}
\label{tab:full-result}
\begin{tabular}{lccc}
\toprule
 & Recover Count $\uparrow$
 & Avg. Steps $\downarrow$
 & Tokens $\downarrow$ \\
\midrule

\rowcolor{gray!15}
\multicolumn{4}{l}{\textbf{Claude Sonnet 5}} \\
\quad Oracle$^\dagger$ & 40/40 (100\%) & 3.03 & 285,417 \\
\quad Naive            & 20/40 (50.0\%) & 4.40 & 589,720 \\
\quad Alert            & 26/40 (65.0\%) & 4.33 & 609,750 \\
\quad Diff             & 36/40 (90.0\%) & 7.10 & 1,079,420 \\
\quad \textbf{Concord}          & \best{40/40 (100\%)} & \best{4.03} & \best{502,596} \\

\rowcolor{gray!15}
\multicolumn{4}{l}{\textbf{GLM 5.2}} \\
\quad Oracle$^\dagger$ & 40/40 (100\%) & 3.70 & 219,753 \\
\quad Naive            & 15/40 (37.5\%) & 4.68 & 408,511 \\
\quad Alert            & 23/40 (57.5\%) & \best{4.65} & 430,712 \\
\quad Diff             & 31/40 (77.5\%) & 6.73 & 663,430 \\
\quad \textbf{Concord}          & \best{40/40 (100\%)} & 4.75 & \best{389,953} \\

\rowcolor{gray!15}
\multicolumn{4}{l}{\textbf{DeepSeek V4 Pro}} \\
\quad Oracle$^\dagger$ & 40/40 (100\%) & 4.58 & 410,820 \\
\quad Naive            & 24/40 (60.0\%) & 5.60 & 562,851 \\
\quad Alert            & 31/40 (77.5\%) & 5.83 & 621,379 \\
\quad Diff             & 38/40 (95.0\%) & 8.20 & 903,032 \\
\quad \textbf{Concord}          & \best{40/40 (100\%)} & \best{5.38} & \best{524,862} \\

\bottomrule
\end{tabular}
\end{table}
\subsection{Results and Analysis}
Table~\ref{tab:full-result} reports recover count, average steps, and token consumption across three frontier models.
Under the constructed patch-and-restore race conditions, Concord also achieves a recover count of 40/40 for each model and uses fewer tokens than Naive, Alert, and Diff in the evaluated runs. 
Concord additionally requires fewer steps than Diff in all three settings.
These results indicate a better observed accuracy-cost tradeoff for runtime-level context repair than for prompt-level recovery on ConcordBench-Pioneer; they should not be interpreted as a workload-independent guarantee of perfect recovery.
To better understand these observed differences, we next conduct a trace-level analysis using the complete GLM 5.2 runs.

\noindent\textbf{Stale evidence persists inside the trajectory.}
Naive recovers only 15/40 tasks in the traced run.
The dominant failure mode is short and direct: the agent reads mutated evidence, forms an answer, and later repeats that answer after the benchmark restores the file.
Many wrong predictions match the mutated value rather than the gold answer, such as a changed number, year, or entity name.
Even when the runtime intercepts premature answers and restores remaining patches (28/40 Naive tasks), the agent receives no signal that its earlier observation is stale, so the restored workspace does not repair the model-visible trajectory.

\noindent\textbf{Prompt-level signals trade accuracy for cost.}
Alert improves the performance from 15/40 to 23/40, but only when the notice leads to re-grounding.
If an alert is followed by a later \texttt{read\_file}, the agent is correct in 17/18 cases; without a later read, it is correct in only 2/8 cases when the alert is visible and 4/14 cases when the alert is only appended around answer interception.
Diff gives a stronger signal by showing an explicit before/after hunk, reaching 31/40.
However, it turns recovery into a tool protocol: Diff uses 269 events and 663k tokens, compared with 148 events and 220k tokens for Oracle, and 21/40 tasks call \texttt{diff} at least twice because of freshness gating.

\noindent\textbf{Runtime-level reconciliation repairs stale observations in the evaluated traces.}
\ccd achieves a recover count of 40/40 in the evaluated GLM 5.2 run while consuming 390k tokens, fewer than Naive, Alert, and Diff in this setting.
It also needs fewer reads and steps than the strongest prompt-level baseline: 1.30 \texttt{read\_file} calls per task and 4.75 average steps, versus 2.25 reads per task and 6.73 steps for Diff.
The reason is that \ccd patches local stale blocks in place, or removes invalid blocks from prompt projection and forces a re-read when drift is larger.
The next model call therefore sees repaired evidence or an explicit refresh requirement, rather than a stale observation plus an advisory warning.

Within ConcordBench-Pioneer, the evaluated methods show a consistent ordering: alerts are inexpensive but incomplete, diffs recover more evaluated cases at higher cost, and runtime context reconciliation matches the oracle recover count while using fewer tokens than the prompt-level race baselines.

\subsection{Limitations and Failure Modes}
Our evaluation covers 40 HotpotQA-derived tasks, local files, a
restricted tool harness, and a controlled patch-and-restore race.
Therefore, the observed 40/40 recover count for each model does not imply
perfect or workload-independent reliability. 
The current prototype
monitors writes routed through \texttt{apply\_patch}; out-of-band
changes require broader change-detection mechanisms, while polling
may miss rapid updates. 
\ccd also relies on correct source
identifiers, versions, and source-to-observation mappings. 
Large
rewrites, ambiguous span relocation, multi-source observations, or
repeated updates may force re-reading, increasing latency and token
cost. 
Moreover, \ccd does not revise summaries, plans, or decisions
derived from stale observations. 
We leave evaluation on broader
workloads, concurrent agents, repeated update cycles, and non-file
resources to future work.
\section{Related Work}
\noindent \textbf{Consistency in Multi-Agent Workspaces.}
Prior work has studied consistency in multi-agent workspaces mainly at the system or runtime level. 
STORM~\cite{liu2026multi} prevents stale writes through local state consistency checks;
CAID~\cite{geng2026effective} isolates agents in separate workspaces and resolves conflicts at merge time; 
and CodeCRDT~\cite{pugachev2025codecrdt} explores convergence for shared code state. 
Other systems manage state through workspace snapshots or explicit runtime boundaries, such as InfiAgent~\cite{yu2026infiagent} and LemonHarness~\cite{ren2026lemonharness}. 
Recent studies also formalize stale-generation and related concurrency anomalies~\cite{khan2026verified}, or show that stale environmental observations can cause grounding failures~\cite{sheng2026agents}. 
These works primarily manage workspace state, write operations, or snapshot reconstruction. 
In contrast, we study prompt-time coherence by tracking observations and handling source changes through invalidation, refresh, or annotation.

\noindent \textbf{Agent Memory Systems.}
This line of work studies how agents manage long-term memory and maintain the freshness of stored facts. 
MemGPT~\cite{packer2023memgpt} introduces an OS-inspired memory hierarchy for managing limited context windows; 
Self-GC~\cite{hao2026self} treats context as indexed objects with lifecycle operations such as folding, masking, and pruning;
and VISTA~\cite{xu2026llm} represents working memory as typed and addressable blocks. 
Recent work also examines memory staleness and temporal validity: 
STALE~\cite{chao2026stale} evaluates whether agents can recognize when prior memories become invalid; 
Governed Shared Memory~\cite{margalit2026governed} studies stale propagation and provenance collapse in shared agent memory; 
and Engram~\cite{wang2026less} and Supersede~\cite{patel2026supersede} use temporal validity or supersession chains to address delayed memory updates. 
These systems are complementary to ours: they mainly focus on retrieval-time evidence selection, memory capacity management, or conversational fact updates, whereas we track tool observations that have already been materialized in an agent's private prompt context.

\noindent \textbf{Cache Coherence in Computer Systems.}
The consistency problem between agent context and external state can be naturally related to cache coherence in classical computer systems. 
From this perspective, mutable external resources, such as file systems or MCP services, serve as the source of truth, while tool observations recorded in an agent's prompt act as cached copies of that state. 
Traditional cache-coherence mechanisms~\cite{adve1996shared, goodman1983using, sorin2011primer} maintain consistency between cached copies and the source state through techniques such as versioning, invalidation, and update propagation.
These ideas provide useful guidance for agent context management, such as maintaining source metadata, dependency information, and invalidation policies for context entries.

\vspace{-5pt}
\section{Conclusion and Future Work}
In this work, we identify and formally define incoherence in volatile agent context.
To address the problem, we propose \ccd, a context coherence framework that maintains the consistency between tool observations in agent context and the mutable external resources.
\ccd is applicable across different agent runtimes and external resources, and can be easily extended to new runtime-resource settings.

Several directions remain for future work. 
First, while our current use case focuses on file-based tool interactions, it can be extended to broader classes of mutable resources (e.g. MCP Tools).
Second, our current implementation focuses on tool observations, but agents often produce summaries, plans, assumptions, or decisions that are derived from those observations. 
Tracking the validity of such derived context is a harder problem and requires richer context dependency modeling. 
Finally, regarding the reconciliation policy, we could explore more efficient strategies, such as finer-grained replacement, batched propagation, or lazy updates, to reduce runtime overhead and avoid unnecessary modifications to the agent context.

\newpage
\bibliographystyle{plain}
\bibliography{references}

@misc{openclaw2026context,
  title        = {{OpenClaw Documentation: Context}},
  author       = {{OpenClaw Contributors}},
  year         = {2026},
  howpublished = {\url{https://docs.openclaw.ai/concepts/context}},
  note         = {Accessed: 2026-07-08}
}

@misc{openclaw2026tools,
  title        = {{OpenClaw Documentation: Tools}},
  author       = {{OpenClaw Contributors}},
  year         = {2026},
  howpublished = {\url{https://github.com/openclaw/openclaw/blob/main/docs/tools/index.md}},
  note         = {Accessed: 2026-07-08}
}

@misc{opencode2026mcp,
  title        = {{OpenCode Documentation: MCP Servers}},
  author       = {{OpenCode Contributors}},
  year         = {2026},
  howpublished = {\url{https://opencode.ai/docs/mcp-servers/}},
  note         = {Accessed: 2026-07-08}
}

@misc{opencode2026github,
  title        = {{OpenCode: A Powerful AI Coding Agent}},
  author       = {{OpenCode Contributors}},
  year         = {2026},
  howpublished = {\url{https://github.com/opencode-ai/opencode}},
  note         = {Accessed: 2026-07-08}
}

@inproceedings{wang2025openhands,
  title={Openhands: An open platform for ai software developers as generalist agents},
  author={Wang, Xingyao and Li, Boxuan and Song, Yufan and Xu, Frank F and Tang, Xiangru and Zhuge, Mingchen and Pan, Jiayi and Song, Yueqi and Li, Bowen and Singh, Jaskirat and others},
  booktitle={International Conference on Learning Representations},
  volume={2025},
  pages={65882--65919},
  year={2025}
}

@misc{anthropic2024mcp,
  title        = {Introducing the Model Context Protocol},
  author       = {{Anthropic}},
  year         = {2024},
  month        = nov,
  howpublished = {\url{https://www.anthropic.com/news/model-context-protocol}},
  note         = {Accessed: 2026-07-08}
}

@inproceedings{hong2024metagpt,
  title={MetaGPT: Meta programming for a multi-agent collaborative framework},
  author={Hong, Sirui and Zhuge, Mingchen and Chen, Jonathan and Zheng, Xiawu and Cheng, Yuheng and Wang, Jinlin and Zhang, Ceyao and Yau, Steven and Lin, Zijuan and Zhou, Liyang and others},
  booktitle={International Conference on Learning Representations},
  volume={2024},
  pages={23247--23275},
  year={2024}
}

@article{liu2026multi,
  title={Multi-agent Collaboration with State Management},
  author={Liu, Mengyang and Chen, Taozhi and Xu, Zhenhua and Jiang, Xue and Dong, Yihong},
  journal={arXiv preprint arXiv:2605.20563},
  year={2026}
}

@article{geng2026effective,
  title={Effective strategies for asynchronous software engineering agents},
  author={Geng, Jiayi and Neubig, Graham},
  journal={arXiv preprint arXiv:2603.21489},
  year={2026}
}

@article{pugachev2025codecrdt,
  title={CodeCRDT: Observation-Driven Coordination for Multi-Agent LLM Code Generation},
  author={Pugachev, Sergey},
  journal={arXiv preprint arXiv:2510.18893},
  year={2025}
}

@article{ren2026lemonharness,
  title={LemonHarness Technical Report},
  author={Ren, Kailong and Sun, Fubo and Liu, Jiachen and Yang, Liu and Yin, Zimo and Li, Jiaying and Yin, Congli and He, Ming and Huo, Yu and Liu, Jiawei and others},
  journal={arXiv preprint arXiv:2606.24311},
  year={2026}
}

@inproceedings{yu2026infiagent,
  title={InfiAgent: An Infinite-Horizon Framework for General-Purpose Autonomous Agents},
  author={Yu, Chenglin and Wang, Yuchen and Wang, Songmiao and Yang, Hongxia and Ming, Li},
  booktitle={Findings of the Association for Computational Linguistics: ACL 2026},
  pages={35884--35894},
  year={2026}
}

@article{hao2026self,
  title={Self-GC: Self-Governing Context for Long-Horizon LLM Agents},
  author={Hao, Xubin and Meng, Hongjin and Yin, Xin and Zhu, Jiawei and Cao, Chenpeng},
  journal={arXiv preprint arXiv:2607.00692},
  year={2026}
}

@article{wang2026conflictrag,
  title={ConflictRAG: Detecting and Resolving Knowledge Conflicts in Retrieval Augmented Generation},
  author={Wang, Chenyu and Li, Yueyuan and Liu, Yingmin and Shu, Yang},
  journal={arXiv preprint arXiv:2605.17301},
  year={2026}
}

@article{margalit2026governed,
  title={Governed Shared Memory for Multi-Agent LLM Systems},
  author={Margalit, Yanki and Cohen-Inger, Nurit and Avram, Erni and Taig, Ran and Margalit, Oded},
  journal={arXiv preprint arXiv:2606.24535},
  year={2026}
}

@misc{postgres_mcp,
  author       = {crystaldba},
  title        = {postgres-mcp: PostgreSQL Model Context Protocol Server},
  howpublished = {\url{https://github.com/crystaldba/postgres-mcp}},
  year         = {2024},
  note         = {Accessed: 2026-07-08}
}

@misc{slack_mcp_server,
  author       = {korotovsky},
  title        = {slack-mcp-server: Slack Model Context Protocol Server},
  howpublished = {\url{https://github.com/korotovsky/slack-mcp-server}},
  year         = {2024},
  note         = {Accessed: 2026-07-08}
}

@misc{github_mcp_server,
  author       = {{GitHub}},
  title        = {github-mcp-server: GitHub Model Context Protocol Server},
  howpublished = {\url{https://github.com/github/github-mcp-server}},
  year         = {2024},
  note         = {Accessed: 2026-07-08}
}

@article{khan2026verified,
  title={Verified Detection and Prevention of Concurrency Anomalies in Multi-Agent Large Language Model Systems},
  author={Khan, Sajjad},
  journal={arXiv preprint arXiv:2606.17182},
  year={2026}
}

@article{sheng2026agents,
  title={When Agents Overtrust Environmental Evidence: An Extensible Agentic Framework for Benchmarking Evidence-Grounding Defects in LLM Agents},
  author={Sheng, Strick and Wang, Ziyue and Zhou, Liyi},
  journal={arXiv preprint arXiv:2605.08828},
  year={2026}
}

@article{packer2023memgpt,
  title={MemGPT: towards LLMs as operating systems.},
  author={Packer, Charles and Fang, Vivian and Patil, Shishir\_G and Lin, Kevin and Wooders, Sarah and Gonzalez, Joseph\_E},
  year={2023},
  publisher={ArXiv}
}

@article{xu2026llm,
  title={LLM Agents Are Latent Context Managers: Eliciting Self-Managed Context via a Proprioceptive Dashboard},
  author={Xu, Binyan and Li, Haitao and Zhang, Kehuan},
  journal={arXiv preprint arXiv:2606.30005},
  year={2026}
}

@article{chao2026stale,
  title={STALE: Can LLM Agents Know When Their Memories Are No Longer Valid?},
  author={Chao, Hanxiang and Bai, Yihan and Sheng, Rui and Li, Tianle and Sun, Yushi},
  journal={arXiv preprint arXiv:2605.06527},
  year={2026}
}

@article{wang2026less,
  title={Less Context, More Accuracy: A Bi-Temporal Memory Engine for LLM Agents Where a Lean Retrieved Context Beats the Full History},
  author={Wang, Liuyin},
  journal={arXiv preprint arXiv:2606.09900},
  year={2026}
}

@article{patel2026supersede,
  title={Supersede: Diagnosing and Training the Memory-Update Gap in LLM Agents},
  author={Patel, Vedant},
  journal={arXiv preprint arXiv:2606.27472},
  year={2026}
}

@inproceedings{goodman1983using,
  title={Using cache memory to reduce processor-memory traffic},
  author={Goodman, James R},
  booktitle={Proceedings of the 10th annual international symposium on Computer architecture},
  pages={124--131},
  year={1983}
}

@book{sorin2011primer,
  title={A primer on memory consistency and cache coherence},
  author={Sorin, Daniel and Hill, Mark and Wood, David},
  year={2011},
  publisher={Morgan \& Claypool Publishers}
}

@article{adve1996shared,
  title={Shared memory consistency models: A tutorial},
  author={Adve, Sarita V and Gharachorloo, Kourosh},
  journal={computer},
  volume={29},
  number={12},
  pages={66--76},
  year={1996},
  publisher={IEEE}
}

@inproceedings{yang2018hotpotqa,
  title={HotpotQA: A dataset for diverse, explainable multi-hop question answering},
  author={Yang, Zhilin and Qi, Peng and Zhang, Saizheng and Bengio, Yoshua and Cohen, William and Salakhutdinov, Ruslan and Manning, Christopher D},
  booktitle={Proceedings of the 2018 conference on empirical methods in natural language processing},
  pages={2369--2380},
  year={2018}
}

\end{document}